\documentclass[a4paper]{article}

\usepackage{INTERSPEECH2018}
\usepackage{amsmath}
\usepackage{amssymb,bm}
\usepackage{textcomp}
\usepackage{url}
\usepackage{multirow}
\usepackage{microtype}
\usepackage{rotating}
\usepackage{subfig}
\def\vec#1{\ensuremath{\bm{{#1}}}}

\title{Fast ASR-free and almost zero-resource keyword spotting using \\ DTW and CNNs for humanitarian monitoring} 
\name{Raghav Menon$^1$, Herman Kamper$^1$, John Quinn$^2$, Thomas Niesler$^1$}
\address{
  $^1$Department of Electrical and Electronic Engineering,  Stellenbosch University, South Africa\\
  $^2$UN Global Pulse, Kampala, Uganda}
\email{rmenon@sun.ac.za, kamperh@sun.ac.za, john.quinn@unglobalpulse.org, trn@sun.ac.za}

\usepackage[prependcaption,textsize=scriptsize]{todonotes}
\usepackage{xcolor}
\definecolor{mycolor}{HTML}{FF6600}

\begin{document}

\maketitle
\begin{abstract}
% Yet another one
We use dynamic time warping (DTW) as supervision for training a convolutional neural network (CNN) based keyword spotting system using a small set of spoken isolated keywords. 
The aim is to allow rapid deployment of a keyword spotting system in a new language to support urgent United Nations (UN) relief programmes in parts of Africa where languages are extremely under-resourced and the development of annotated speech resources is infeasible.
First, we use 1920 recorded keywords (40 keyword types, 34 minutes of speech) as exemplars in a DTW-based template matching system and apply it to untranscribed broadcast speech.
Then, we use the resulting DTW scores as targets to train a CNN on the same unlabelled speech.
In this way we use just 34 minutes of labelled speech, but leverage a large amount of unlabelled data for training.
While the resulting CNN keyword spotter cannot match the performance of the DTW-based system, it substantially outperforms a CNN classifier trained only on the keywords, improving the area under the ROC curve from 0.54 to 0.64.
Because our CNN system is several orders of magnitude faster at runtime than the DTW system, it represents the most viable keyword spotter on this extremely limited dataset.

% We use dynamic time warping (DTW) as supervision for training a convolutional neural network (CNN) based keyword spotting system using a very small set of spoken isolated keywords.
% The aim is to allow rapid deployment of a keyword spotting system in a new language to support urgent United Nations (UN) relief programmes in parts of Africa where languages are extremely under-resourced and the development of annotated speech resources is infeasible.
% First, we use the \textcolor{red}{[trn: say how many KWs]} recorded keywords (34 minutes of speech) as exemplars in a DTW-based template matching system and apply it to untranscribed broadcast speech.
% Then, we use the resulting DTW scores as targets to train a CNN on the \textcolor{red}{same (?)} unlabelled speech.
% In this way we use just 34 minutes of labelled data leverage a much larger amount of unlabelled data for training.
% We show that, while the resulting CNN keyword spotter cannot match the performance of the DTW-based system, it substantially outperforms a CNN classifier trained only on the isolated keywords, improving the area under the ROC curve (AUC) from 0.54 to 0.64.
% Because our CNN system is several orders of magnitude faster at runtime than the DTW system, it represents the most viable keyword spotter on this extremely limited dataset.

\end{abstract}


\section{Introduction}

In societies with well-developed internet infrastructure, social media has become a dominant medium for voicing views and concerns about various social issues~\cite{Vosoughi_ICDMW15,Wegrzyn_CASoN11,Burnap15}.
% affecting  society
In countries like Uganda, where internet availability is limited, phone-in talk shows on local community radio stations are used in a similar way.
A United Nations (UN) piloted project has developed radio-browsing systems for monitoring such radio shows in order to inform relief and developmental purposes. 
These systems have been very successful and are in active use.
% A \herman{United Nations} (UN) piloted project for monitoring the chat shows, called radio-browsing, and its deployment for relief and developmental purpose has been a huge success.
% Social media has been a prevalent mode of communication in societies with sufficiently good internet infrastructure. They have been used quite often by the population to put forward their concerns on various issues affecting the society \cite{Vosoughi_ICDMW15,Wegrzyn_CASoN11,Burnap15}. In countries like Uganda where such an infrastructure is poor, people make use of phone in talk shows hosted by the local community radio stations in order to voice the same. A United Nations (UN) piloted project for monitoring the transmitted chat shows, called radio-browsing, and its deployment for relief and developmental purpose has been a huge success. 

In previous work~\cite{Menon2017,Saeb2017}, we assumed the availability of small amounts of transcribed data (initially about 9 hours and then just 12 minutes) for the development and deployment of such systems in two regional languages, Luganda and Acholi. However, even the preparation of a 12 minute transcribed corpus requires the availability of annotators in the new language with appropriate skills. 
This has proved a serious impediment to the development of a radio browsing system in a new language.
Here we therefore turn our attention to the development of a %n ASR-free
keyword spotter that can be set up using resources that are even easier to obtain: a small set of isolated spoken keywords.

Recent advances in automatic speech recognition (ASR) technology have mostly been restricted to scenarios where very large transcribed speech resources are available~\cite{Sainath2015,Zhang2017}. 
For keyword spotting, where the goal is to search a speech signal for occurrences of a keyword provided as text, most systems employ ASR to produce lattices which are subsequently searched~\cite{Larson12, Mandal14}. 
This is not feasible without at least a minimal corpus of transcribed speech in the target domain. 

%In many settings, complete transcribed data is not available for developing ASR systems which affects its performance. \todo{HK: Need to say in one sentence what keyword spotting is. Something like: ``For keyword spotting, where the goal is to search for a given text keyword in speech collection, most systems employ an ASR recogniser to produce lattices as a first stage. In many settings, complete transcribed data is not available for developing ASR systems.'' } \todo[color=green!40]{RM:Done}
% Major developments in the area of speech recognition and its applications over the last few years have mostly been restricted to scenarios where large transcribed speech resources are available \cite{Zhang2017, Sainath2015}. Keyword spotting \todo{HK: Need to say in one sentence what keyword spotting is. Something like: ``For keyword spotting, where the goal is to search for a given text keyword in speech collection, most systems employ an ASR recogniser to produce lattices as a first stage. In many settings, complete transcribed data is not available for developing ASR systems.'' } using speech has mostly been dependent on the performance of ASR since the first stage for such an application requires the speech to be recognized \cite{Larson12, Mandal14}.

In settings where transcribed data is not available, researchers have attempted to achieve ASR-free keyword spotting by adopting a query-by-example (QbyE) retrieval procedure. 
%\todo{HK: Very often, they simply just use ``QbE'' as the acronym.}
In QbyE, the search query is provided not as text but as audio.
% These systems typically use 
Typically, dynamic time warping (DTW) is used to match the acoustic features from the search query to acoustic features from speech in the search collection~\cite{Hazen2009, Zhang2009}. 
Alternatively, queries and search utterances can be mapped into a joint fixed-dimensional space~\cite{Levin2013,Chung2016}, allowing for efficient retrieval using vector comparisons.
% These technoques do not rely on neural networks and are specifically aimed at QbyE in a low-resource setting. 
Such fixed-dimensional vector representations can be obtained using recurrent neural networks~\cite{Chen2015,settle+livescu_slt16} or, when matching word pairs are known, a Siamese convolutional neural network (CNN)~\cite{Kamper2016}.
%\todo{HK: I know this is a LSTM, but I don't think this is directly important} %LSTM
%\todo{HK: To answer TRN's question: these are also trained to learn vector representations (not to predict distances). They are really trained to predict same vs.\ different labels.}
% The network is applied on a window that slides over the input speech in order to spot a keyword using the cosine distance.
% after which the keyword search proceeds with the help of sliding window and the cosine distance. 
% \textcolor{red}{[trn: have I broken the previous sentence?]}
 % Also here the cosine distance is used to compute the similarity between the search query and speech segments.
% A Siamese convolutional neural network (CNN) can also be trained to determine the similarity scores between the target speech and the word for keyword spotting~\cite{Kamper2016}. 
% \textcolor{red}{[trn: does the SCNN compute the distance or also a vector representation?]}
In \cite{Audhkhasi2017}, an ASR-free keyword spotter is described which maps textual and acoustic input into a shared fixed-dimensional space where text queries can be compared directly to search utterances.
% which inputs the search query in the form of text: \herman{a distance is calculated between a textual word embedding and a fixed-dimensional acoustic embedding.}
% \textcolor{red}{[trn: more detail needed here - all the methods discussed in this par are ASR-free. How does this one work?]}
However, all these ASR-free neural QbyE approaches rely on large amounts of training data. 
 
In this paper, we combine DTW and CNNs to develop an ASR-free keyword spotter that is trained on an easy-to-obtain small number of isolated keyword utterances.
Specifically, we use a small seed corpus and DTW to calculate training targets for a much larger unannotated corpus, that can subsequently be used to train a CNN-based keyword spotter.
The CNN model is much faster than the DTW-based system (since alignment is not required), making this a viable option for real-time monitoring.

% \herman{A major benefit of using the CNN instead of the DTW-based system is }
%The main intention is to investigate the possibility of building a keyword spotter which is ASR-free and when there is almost no training data available. 
%Due to the presence of human analysts in the final stage of the radio-browsing system, it is possible for us to compromise on the false positive rate in order to ensure that the target keywords are not missed.

%The paper is organised as follows: Section~\ref{Sec_Radio} describes the radio-browsing system with our suggested modification. Section~\ref{Sec_Data} describes the data and section 4 goes around describing the system. Section~\ref{Sec:Exp} presents the experimental conducted and discusses the results with section~\ref{Sec:CONclusion} concluding the paper.\todo{HK: If space is a problem, this paragraph can be cut.}

\section{Radio Browsing System Configuration}
\label{Sec_Radio}
\begin{figure}[t]
  \centering
  \captionsetup{justification=centering}
  \includegraphics[width=0.95\linewidth]{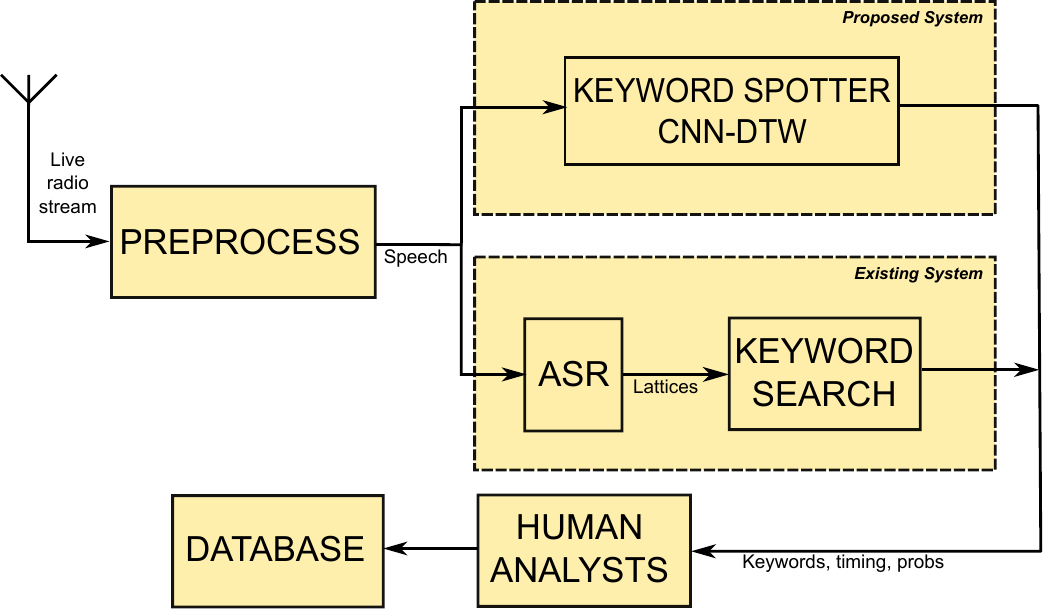}
  \caption{Radio browsing system showing the existing and the proposed system.}
  \vspace{-12pt}
  \label{fig:radio-browsing}
\end{figure}

To give the broader context for our work, we first describe our complete radio browsing system, shown in Figure~\ref{fig:radio-browsing}.
This figure also shows our previous ASR-based implementation, in which pre-processed speech from a live audio stream is passed to an ASR system which generates lattices.
These lattices are then indexed and searched for the desired keywords. 
Our new system replaces the ASR components with a CNN-DTW keyword spotter.
The detected keywords and their meta-data are passed to human analysts who filter the information and format it into a structured, categorised and searchable format appropriate for humanitarian decision making and situational awareness. 
In this scenario, high false positive rates can be tolerated because the human analysts can discard false detections. 
The overall approach allows the analysis of a large amount of audio data while maintaining a high confidence in the final output. 
A more detailed discussion on the role of human analysts and  examples of detected topics of interest are presented in~\cite{Saeb2017}.\footnote{Examples available at~\mbox{\url{http://radio.unglobalpulse.net}.}}

\section{Data}
\label{Sec_Data}
In this first work we use a corpus of South African Broadcast News (SABN) for experimental analysis. 
Since transcriptions are available for this data, it allows system performance to be experimentally evaluated.
However in all other respects we consider the SABN data as untranscribed.
Ultimately our goal is to apply this system to languages such as Somali, Rutooro and Lugbara, for which no language resources are available at all. 

The SABN corpus consists of 23 hours of speech from news bulletins broadcast between 1996 and 2006 by one of the {South Africa's} main radio news channels, SAFM~\cite{Kamper2015}. 
The corpus contains a mix of newsreader speech, interviews, and crossings to reporters. 
About 80\% of the speakers can be considered native English speakers.
The division of the corpus into training, development and test sets is shown in Table~\ref{SABC_data}. 

\begin{table}[h]
\centering
\vspace*{-5pt}
\caption{The South African Broadcast News (SABN) dataset.}
\vspace*{-7.5pt}
\label{SABC_data}
{\eightpt
\renewcommand{\arraystretch}{1.2}
\begin{tabular}{|l|c|c|}
\hline
               & \textbf{Utterances} & \textbf{Speech (h)} \\ \hline\hline
\textbf{Train} & 5231                & 7.94                \\ 
\textbf{Dev}   & 2988                & 5.37                \\ 
\textbf{Test}  & 5226                & 10.33               \\ \hline
\textbf{Total} & 13445               & 23.64               \\ \hline
\end{tabular}}
\end{table}

% \begin{figure}[h]
%   \centering
%   \includegraphics[width=\linewidth]{Keyword_distribution_corpus.pdf}
%   \caption{Keyword Distribution in SABN Corpus.}
%   \vspace{-14pt}
%   \label{fig:key_dist}
% \end{figure}

For the purpose of training our keyword spotter, we recorded isolated utterances of 40 keywords, each spoken twice by 24 speakers (12 male and 12 female), leading to a set of 1920 labeled isolated keywords. 
Keywords were displayed to the speakers sequentially, individually and in no particular order.
Keywords were recorded in a quiet room under very different audio conditions and by different speakers than those in the SABN data.
This is representative of the intended operational setting of our system, where isolated utterances of the required keywords will be recorded on location from native speakers using smart phones or audio recorders in conditions that will differ from the radio broadcasts that must be~monitored.

%The distribution of the occurrence of the target keywords within the SABN {\it{train}}, {\it{dev}} and {\it{test}} sets are shown in Figure~\ref{fig:key_dist}. 
%We see that the keyword distribution is skewed over the three sets where in the number of keywords in {\it{train}} set to train on is very sparse.

\section{Keyword Spotting}
\label{Sec_kyproposed}

Here we describe our proposed approach of using DTW {and a small set of individually recorded keywords to train a CNN on a a much larger, untranscribed corpus.}
We briefly describe DTW- and CNN-based keyword spotting approaches before describing our combined procedure.

\subsection{Keyword spotting using dynamic time warping}

When only few isolated spoken keywords are available, dynamic time warping (DTW) is an appropriate technique for their detection since this technique requires as little as a single audio template. 
DTW aligns two sequences of feature vectors by warping their time axes to achieve an optimal match. 
The associated alignment cost can be used as a metric of similarity between sequences.
For keyword spotting with DTW, features are extracted for both the keyword and the search utterances.
To determine whether a keyword is present in an utterance, a naive method is to compute the DTW cost of aligning the keyword with the speech in a sliding window over the search utterance.
More advanced approaches that find subsequences have been proposed~\cite{park+glass_taslp08,jansen+vandurme_interspeech12}, however in this work we restrict ourselves to this simpler implementation. 
We use cosine distance for frame-wise comparison, normalize to obtain per-frame cost, and use a frame-skip of 3 frames.
% The keyword is aligned with successive portions of the utterance, in each case computing the similarity score associated with the best warping of the time axes. 
% We use the cosine similarity as distance metric:
% \begin{equation}
%   d_{cos}(X,Y) = 1-\frac{\sum{x_iy_i}}{\sum{x_i}\sum{y_i}}
%   \label{distance}
% \end{equation}
% where $d_{cos}(X,Y) \in [0,2]$. Thus the similarity measure in Equation~\ref{distance}
The resulting cost $c \in [0, 2]$ has a value of $0$ when the keyword matches a portion of the  utterance exactly and a value of $2$ when they are dissimilar. 
%\todo{HK: I removed the definition of cosine distance, since it's easy to look up and took up quite a lot of space.}
% The result of this process is a time series of DTW similarities indicating how well the keyword matches the speech being searched at every considered point in time. 
By choosing an appropriate threshold for $c$, it is possible to take a decision regarding the presence of the keyword in unlabeled speech.
Although useful in low-resource settings, a major disadvantage of DTW-based search is that it requires alignment between keywords and search utterances, which can be prohibitively slow.
% (depending on the number of keywords and the size of the search collection)

\subsection{Convolutional neural network keyword spotting}
\label{sec:cnn}
In fully supervised settings where a large number of labelled keywords are available, an end-to-end keyword spotter could be trained to directly classify whether a keyword is present in a search utterance.
Although we have only a limited set of labelled spoken keyword, we nevertheless attempt to train a supervised CNN classifier in this way.
%\todo{HK: I also removed the CNN description, since I think most people will be familiar with this. But the paragraph is in the comments if you think we should add it back in.}
% A typical convolutional neural network (CNN) consist of inputs connected to one or more convolutional layers which can be followed by optional max-pool layers and then fully connected layers. 
% The convolutional layer consists of a group of filters which slides over the time sequence of feature vectors extracted from the speech.
% The output of the convolutional layer is an element-wise multiplication of the filter with the area under it called the receptive field. 
% Thus the convolutional layer applies the same transformation to all the speech frames and the number of transformations applied depends on the number of filters in the convolutional layer.  
% The optional max-pooling layer, which usually follows the convolutional layer, is a sub-sampling layer which retains portions of output from the convolutional layer that are important.
% The output of each max-pooling layer forms a feature map.
% These feature maps are then input of a fully connected layer which achieves classification. 
% As a baseline, a CNN was 
This baseline CNN classifier is trained directly in a supervised fashion on the 1920 recorded isolated keywords, as well as negative samples drawn randomly from utterances in the SABN training set.
% which are known not to contain the keywords in question. 
% The input to the CNN was fixed at 60 frames.
% Shorter and longer keywords were adjust by interpolation. 
% \textcolor{red}{[trn: Can we include a citation here - who else has trained a CNN in (approximately) this way, using more data? If it is \cite{SainathPara2015} then ok and make it aparrent.]}
% Keyword spotters using CNNs have been bulit in this way before, but using much larger training sets and a limited number of keywords~\cite{SainathPara2015}.
At test time, a sliding window is applied and keyword presence is predicted based on a threshold. 
Here we used a fixed window of 60 frames.
Similar QbyE and keyword spotting systems based on neural networks have been developed using much larger labeled datasets in previous work~\cite{Kamper2016,palaz+etal_interspeech16,SainathPara2015}.
% After training is done, the test data is presented to the CNN by selecting 60 frames at a time and determining whether the keyword is present or not using a threshold. 

\subsection{CNN-DTW keyword spotting}
\label{Sec:CNN-DTW}
\begin{figure}[t]
  \centering
	\captionsetup{justification=centering}
  \includegraphics[width=\linewidth]{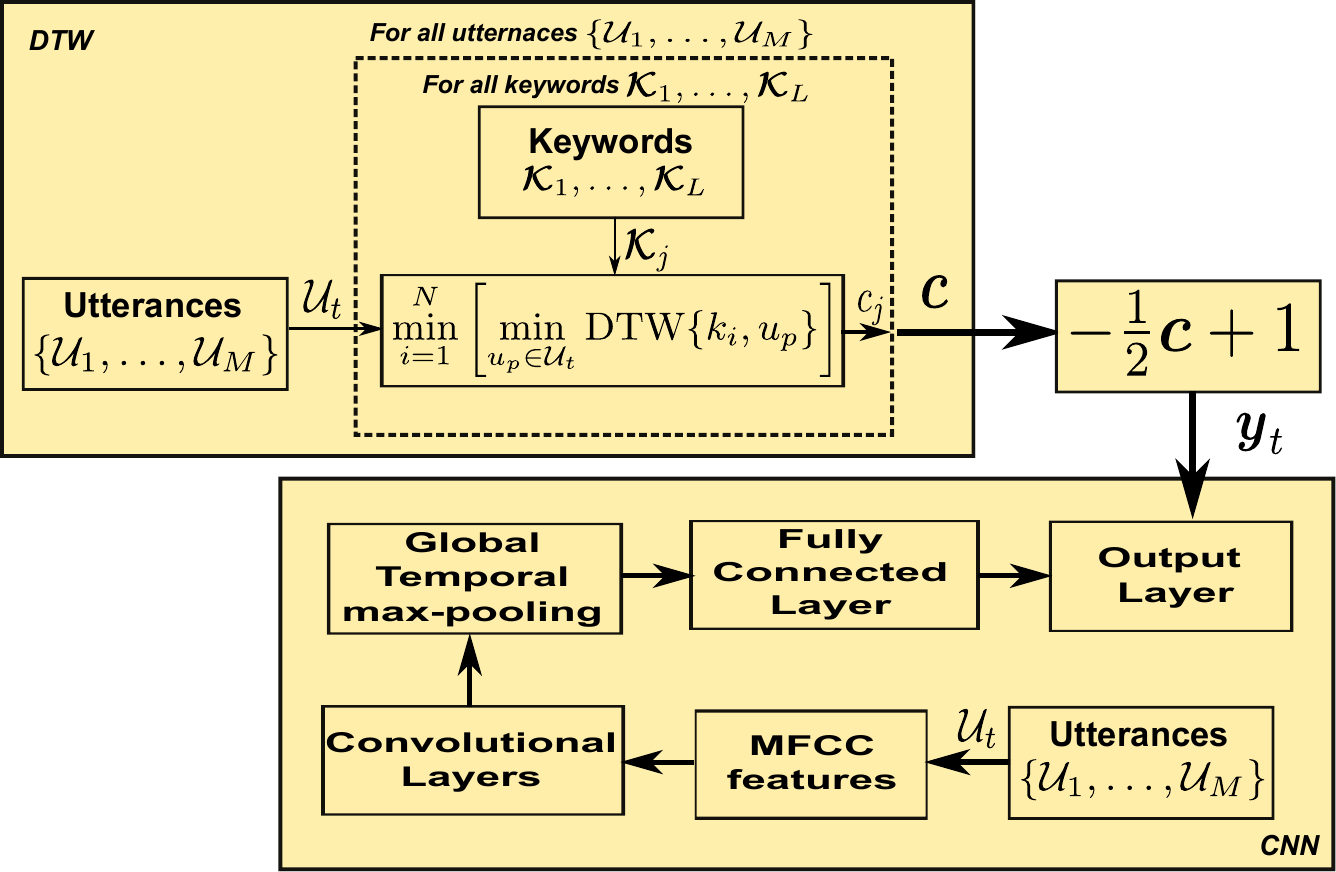}
  \caption{The CNN-DTW keyword spotter training approach. The top shows how the supervisory signal is obtained and the bottom how this signal is used to train the CNN.}
  \vspace{-15pt}
  \label{fig:CNN_DTW}
\end{figure}

CNNs require large amounts of data for training, but are computationally efficient to apply. 
DTW-based keyword spotting, on the other hand, can be applied with only a few keyword exemplars, but is computationally costly.
Our proposal is to employ DTW during training to address the  challenge of data scarcity while taking advantage of the speed benefits of CNNs at runtime.  
% in order to obtain more target data with which a CNN keyword spotter can be trained, 
We achieve this by using DTW to calculate similarity scores between our small set of isolated keywords and a much larger untranscribed dataset and then use this set of similarity scores  as targets to train a CNN.
This strategy is shown in Figure~\ref{fig:CNN_DTW}.
In the upper half of the figure, each repetition of a keyword type (48 in our case) is aligned with an utterance from the untranscribed data using DTW.  
Subsequently, the lowest cost among the 48 repetitions is determined.
This procedure is repeated for all keyword types (40 in our case).  
The result is a vector of scores for each utterance in the untranscribed corpus.
Each dimension of this vector gives an indication of whether the corresponding keyword is present in the particular utterance.
These scores are the targets used to train the CNN, as shown in the lower half of Figure~\ref{fig:CNN_DTW}.
The overall approach therefore relies only on a small set of labeled keywords and a large corpus of untranscribed speech.
% \herman{We now state this process formally.}
% a vector of scores whose dimensionality is the number of different keywords.
% This is carried out for each utterance in the untranscribed corpus.
 
%The result, for a speech utterance, would be a vector of values whose size would be equal to the number of keywords. This resultant vector serves as the target vector for the particular utterance. When this procedure is repeated for all the available speech utterances in the {\it{train}} set we get a matrix of target vectors thus providing the target on which each speech utterances in the {\it{train}} set can be trained. A similar matrix for the {\it{dev}} set is obtained on which the CNN is optimized. %The procedure for determining the target vector is shown using the equations~\ref{per_tar_vec}-\ref{tar_vec} and is also shown in {\it{Block 1}} of Figure~\ref{fig:CNN_DTW}. 

We now state our approach more formally.
Consider a keyword type $\boldsymbol{\mathcal{K}}$ of which we have $N$ repetitions:
\begin{equation}
    \boldsymbol{\mathcal{K}} = (k_{1},\ldots ,k_{i},\ldots, k_{N}) %, l = 1,\ldots, L
    \label{eq:key_vec}
\end{equation}
where each $k_{i}$ is the sequence of speech features for the $i^{th}$ exemplar of keyword $\boldsymbol{\mathcal{K}}$.
To obtain the DTW-based score indicating how likely it is that a particular utterance $\mathcal{U}$ contains an instance of keyword $\boldsymbol{\mathcal{K}}$, we calculate:
 \begin{equation}
    c = \min_{i \in 1 \ldots N} \left[ \min_{u_p \in \mathcal{U}} \textrm{DTW}\{k_{i},{u}_p\} \right]
    \label{eq:tar_vec}
  \end{equation}
Each ${u}_p$ is a successive segment of utterance $\mathcal{U}$, and $\textrm{DTW} \{ k_{i},{u}_p \}$ is the DTW alignment cost between the speech features of exemplar $k_i$ and the segment ${u}_p$.
% where $\textrm{DTW} \{ k_{i},{u}_l \}$ is the DTW alignment cost between the speech features of exemplar $k_i$ and those of a portion $u_l$ of utterance $u$.
Thus, we determine the relevance of a keyword according to the lowest cost encountered when sweeping each of the exemplars over the current utterance.
Assuming we have $L$ keyword types, we calculate equation~\eqref{eq:tar_vec} separately for each of the $L$ keywords.
Hence for utterance $\mathcal{U}$ we obtain costs $[ c_1, \ldots, c_j, \ldots c_L ]$. % for $j = 1, \ldots, L$.
Since we use the averaged cosine distance in our DTW implementation, the costs $c_j \in [0, 2]$.  
In order to interpret these costs as probabilities, we apply the normalization $y_j = -\frac{1}{2}c_j + 1$ so that $y_j \in [0, 1]$, with $1$ indicating a perfect match and $0$ indicating maximum dissimilarity.
Finally, we combine all the normalized scores into a single target vector $\boldsymbol{y} = [ y_1, \ldots, y_L ]$ for utterance $\mathcal{U}$.\footnote{We also considered applying a threshold to obtain hard targets ($1$ indicating the presence and $0$ indicating the absence of the keyword), but this did not improve performance.}

Given this target vector, we train a CNN to take   $\mathcal{U}$ as input and predict this target vector, as illustrated in the lower part of Figure~\ref{fig:CNN_DTW}.
Our CNN consists of a number of convolutional layers, a global temporal max-pooling layer, and a number of fully connected layers.
The global max-pooling layer takes the maximum of the activations over the time dimension, and therefore gives a fixed-dimensional output independent of the length of the input sequence. % \textcolor{red}{(i.e. global-max pooling converts the outputs from the convolutional layer into fixed dimensional vector before the fully connected layer rather than flattening the outputs.)}. 
The intuition is that this would extract the dominant features that are necessary for 
detecting the presence of a keyword in
%predicting the target scores over 
the utterance. 
We use leaky ReLU ($\alpha = 1/3$) for all activations~\cite{xu2015empirical} except the final feedforward layer which uses a sigmoid activation. %\todo{HK: Raghav, add citation to leaky ReLU.} % for score prediction.
Since our scores are normalized to resemble probabilities, we train using the summed binary cross-entropy loss:
\vspace{-5pt}
\begin{align}
%     \mathcal{L}
%     &= -\sum_{j=1}^{L}\sum_{i = 1}^N \left\{ y_{ji} \log \hat{y}_{ji} \;\; +\right. \nonumber \\ 
%     &\qquad\qquad \left. (1 - y_{ji}) \log\left[1 - \hat{y}_{ji} \right] \right\}  %\nonumber \\
	\ell
    &= -\sum_{j = 1}^L \left\{ y_{j} \log \hat{y}_{j} + 
    (1 - y_{j}) \log\left[1 - \hat{y}_{j} \right] \right\}
    \label{eq:binary_cross_entropy}
\end{align}
This is the loss for a single training utterance, with $\hat{y}_j$ the prediction of our model for the $j^\textrm{th}$ keyword type (this would be a single dimension from the model output).
We sum this loss over all $M$ 
%training examples of utterances. 
training utterances in the untranscribed corpus.
Our CNN model can thus be interpreted as $L$ binary classifiers, one for each keyword, with shared input layers.
Finally, the trained CNN can be applied to unseen test utterances, using an appropriate threshold to determined the presence or absence of a keyword.

\vspace{-8pt}

\section{Experiments}
\label{Sec:Exp}
\subsection{Experimental setup}
%The problem we have been facing in the radio-browsing project is thus the availability of large amounts of untranscribed data in many under-resourced languages with little or no experts to annotate the data. Since this project has been for humanitarian and developmental causes, even with the availability of experts it would take quite a lot of time to annotate the data by which the need for such a data would not be existing. We had very little amount of independent recordings of the target keywords whose occurrence had to be detected in the data. 
Three baseline systems are used for comparative evaluation.
Firstly, DTW is performed for each exemplar of a keyword and the resulting scores averaged.
This corresponds to an established QbyE method and will therefore be indicated as DTW-QbyE.
Second, the minimum (best) score over all exemplars of a keyword is used instead of the average.
This will be referred to as DTW-KS.
%Experiments have been performed using DTW which is used as the baseline for the performance comparison. 
%The performance of DTW has been analysed in two configurations 1) By averaging the scores obtained as a result of passing the exemplars of all the keywords over each utterance and then determining the performance over all the utterances (DTW (QbyE) and 2) By choosing the best score for each keyword when passed over each utterance and computing the performance (DTW-KS). 
%\textcolor{red}{[trn: motivate why you are presenting results for these 2 cases. For example, is one close to a standard QbyE setup?]}
Third, we consider the direct application of a CNN classifier trained only on the isolated words, as described in Section~\ref{sec:cnn}.
These three baseline systems are compared with our proposed CNN-DTW approach.

%For the proposed algorithm, target vectors for the {\it{train}} and {\it{dev}} utterances are obtained using the procedure described in Section~\ref{Sec_kyproposed}. These scores which are between $[0,2]$ with $0$ indicating maximum similarity are normalised between $[0,1]$ having $1$ to indicate maximum similarity. \todo{HK: Can move this sentence to where you describe loss; again use math Latex everywhere, i.e. $[0,2]$.} These form the target scores for each utterance which is then trained using CNN having an architecture as shown in {\it{Block 2}} of Figure~\ref{fig:CNN_DTW}. 
The CNN's parameters such as the learning rate, number of convolutional and fully connected layers, number of filters in the convolutional layers, number of neurons in the fully connected layers and the dropout probability for regularization were optimized on the development set;
this optimisation was performed in terms of the summed binary cross-entropy loss of the DTW targets on the development set, meaning that transcriptions of the SABN data are not used for either training of validation.\footnote{Final model: 10 convolutional layers with between 80 and 512 filters per layer, two 3000-unit fully connected layers trained with a dropout of $0.5$, and the learning rate is linearly changed from $10^{-4}$ to $10^{-5}$.} %\todo{HK: Raghav, check this sentence.}
We train using the Adam optimiser~\cite{kingma2014adam} along with an early stopping criterion. %\todo{HK: Raghav}
% transcriptions of the development data are never used.
% to achieve best performance.
% Development optimisation is performed in terms of the loss
% \textcolor{green}{The target values for the {\it{dev}} set is obtained in the same way using the procedure in Section~\ref{Sec:CNN-DTW}.}
% \textcolor{red}{[trn: Instert text herer to make clar that we are not using the SABN transcriptions for this optimisation.]}
Performance is reported in terms of the area under the curve (AUC) of the receiver operating characteristic (ROC). The ROC is a plot of the false positive rate against the true positive rate, as the detection threshold is varied. AUC therefore indicates the performance of the model independent of a threshold, with higher AUC indicating a better model.
We also report equal error rate (EER), the point at which the false positive rate equals the false negative rate (thus, lower EER is better).% \todo{HK: Also just check.}
% The EER is a point where the false positive rate is equal to the false negative rate and hence lower the EER better the model.
% % The threshold for the analysis of our model is set at the point of Equal Error Rate (EER) of the  development set.

\subsection{Results and Discussions}

Table~\ref{Tab:Res1} shows the performance for the different systems in terms of AUC and EER averaged over the 40 keywords. 
From Table~\ref{Tab:Res1} we find that the best result is obtained by the DTW-KS configuration which uses the best scores obtained among all exemplars of a keyword.
The performance of our proposed algorithm combining CNN and DTW is close to the average result of the DTW-QbyE approach.  
%\todo{HK: I don't think the QbE results are really comparable to any of the other scores since they are essentially calculated on different sets? trn: Can this be clarified? It sounds to me that the QbyE is tested on the same set as the other systems ...} 
The performance of the CNN classifier trained only on isolated words is much worse.
% and indicates that the output for most of the keyword is a random guess. 
We found that including a Gaussian noise layer (GNL) between the input and the convolution layers improved the AUC by approximately $1\%$ absolute, showing that this noise layer aids in generalisation. 
% The reason for this improvement may be due to the fact that the perturbation of the training data using noise aids in generalization.  
%This could be an indication that if the training data is augmented by adding gaussian noise then we could further improve the results which is something to be tried in out future.

\begin{figure}[!t]
%  \vspace{-6pt}
  \centering
  \includegraphics[width=\linewidth]{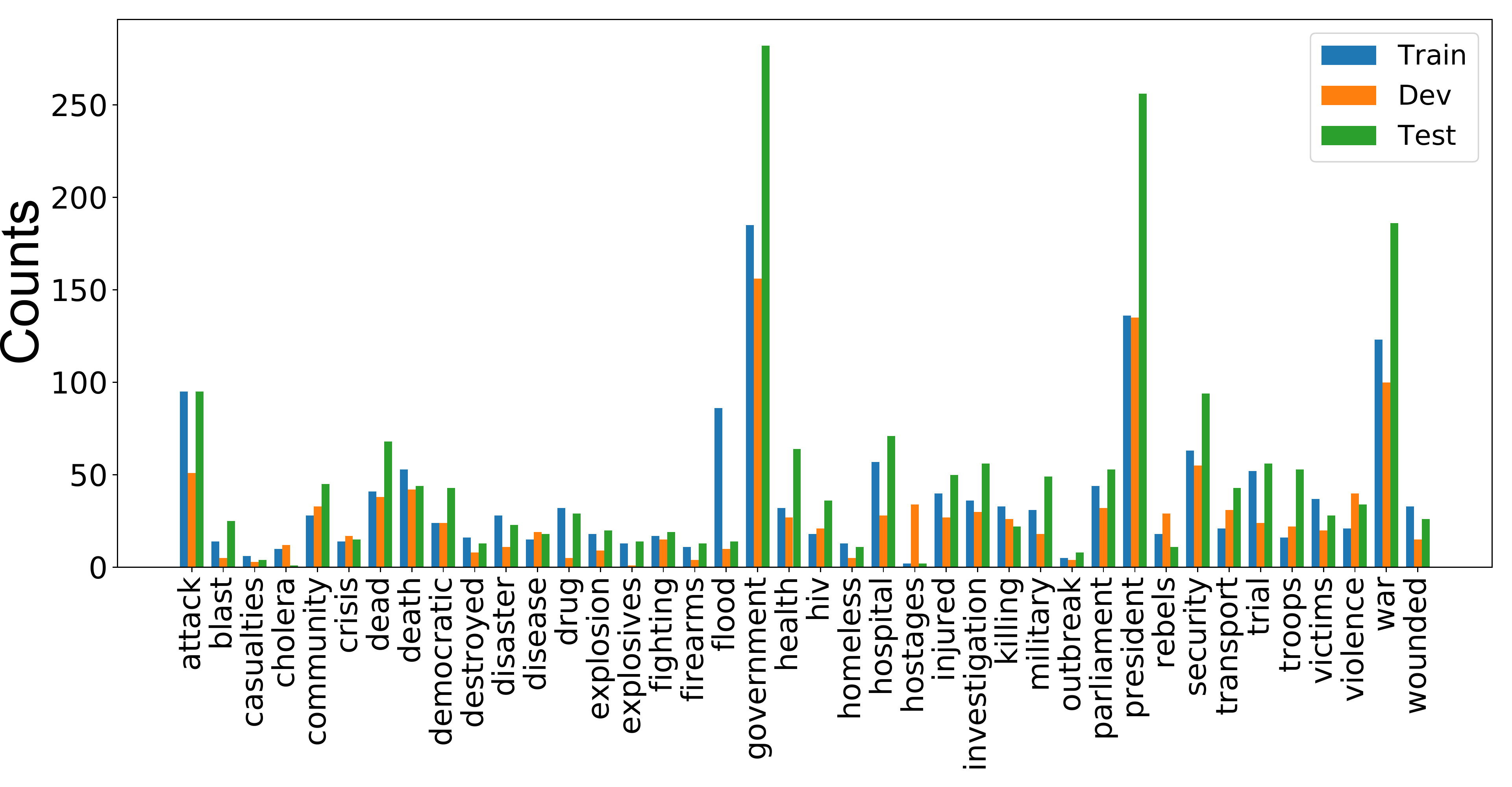}
  \vspace{-20pt}
  \caption{Keyword occurrence distribution in the SABN corpus.}
  \label{fig:key_dist}
\end{figure}

\begin{figure}[!t]
\centering
\captionsetup{justification=centering}
\subfloat[Part 1][\vspace{-4pt}Keyword: Government]{\includegraphics[width=1.5in]{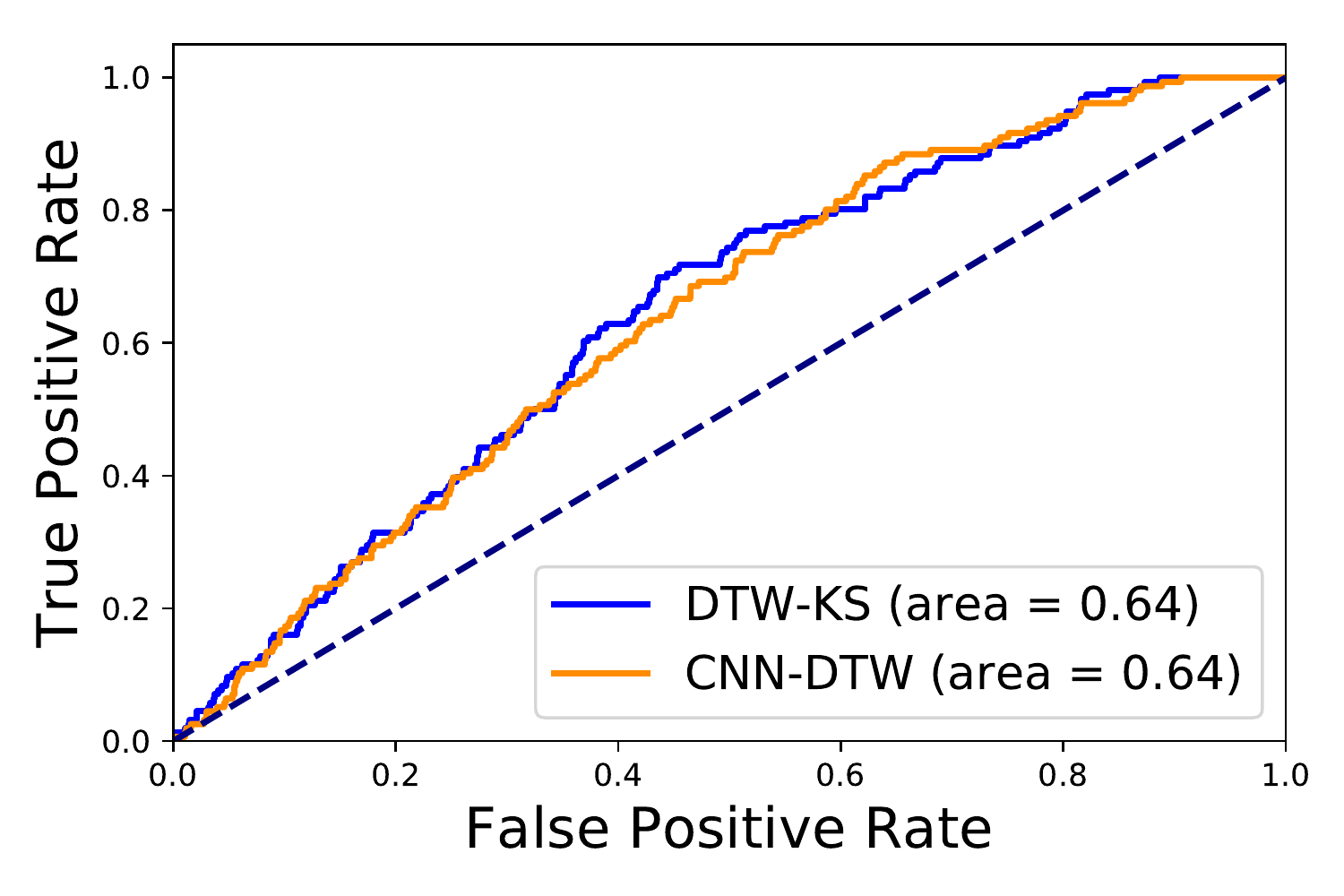} \label{fig:roc_a}}
\subfloat[Part 2][\vspace{-4pt}Keyword: Attack]{\includegraphics[width=1.5in]{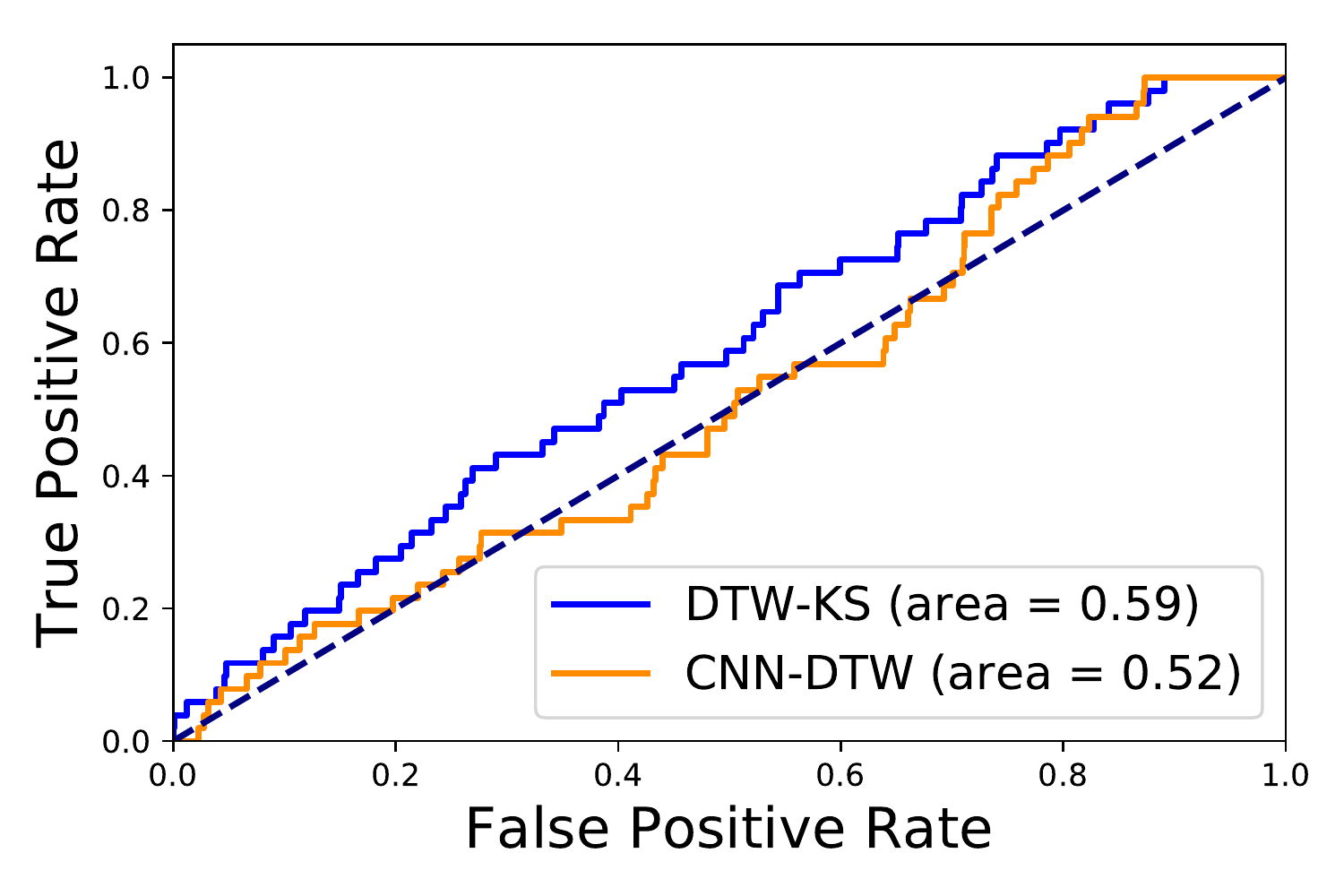} \label{fig:roc_b}}\\[-6pt]
\subfloat[Part 3][\vspace{-4pt}Keyword: HIV]{\includegraphics[width=1.5in]{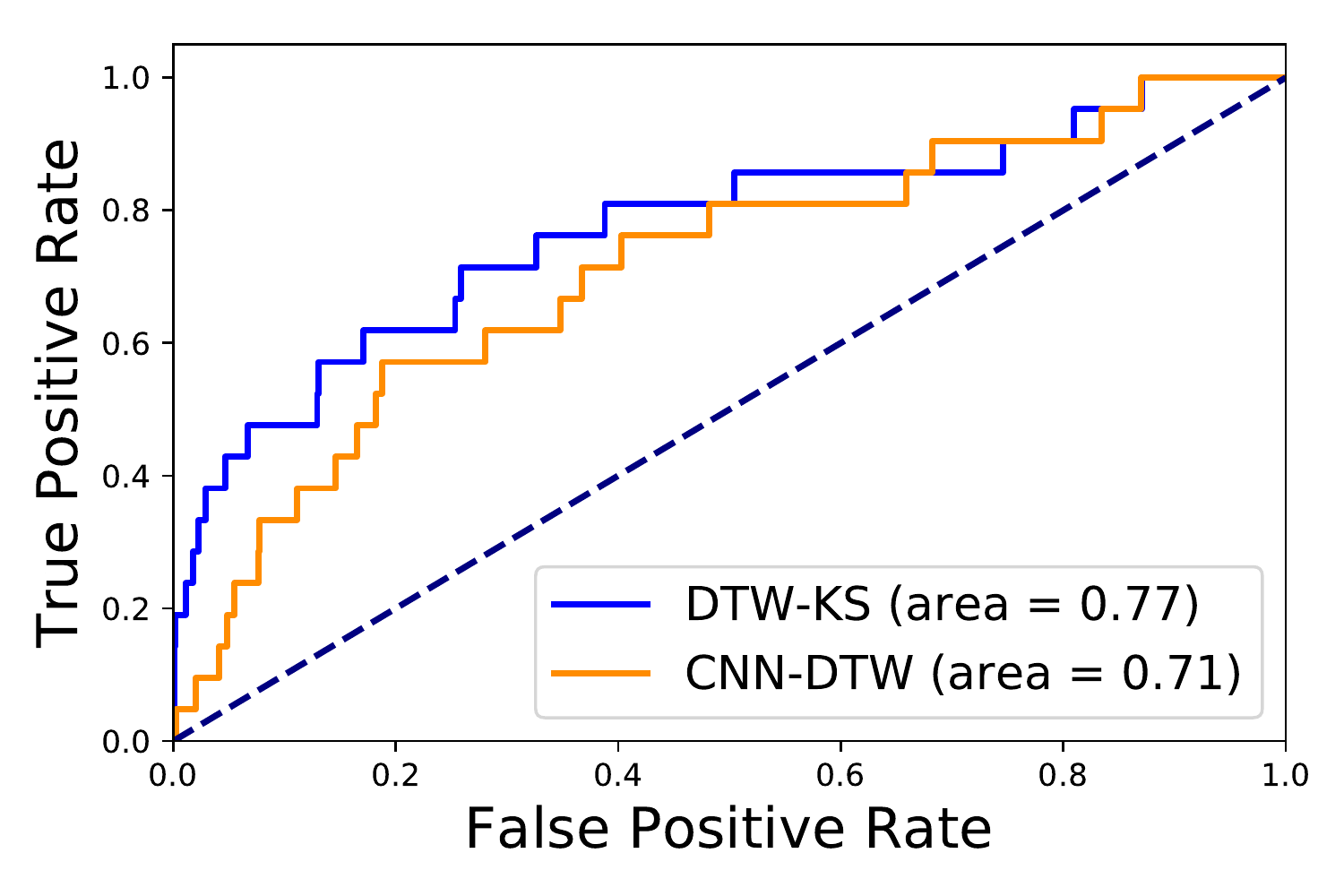} \label{fig:roc_c}}
\subfloat[Part 4][\vspace{-4pt}Keyword: Health]{\includegraphics[width=1.5in]{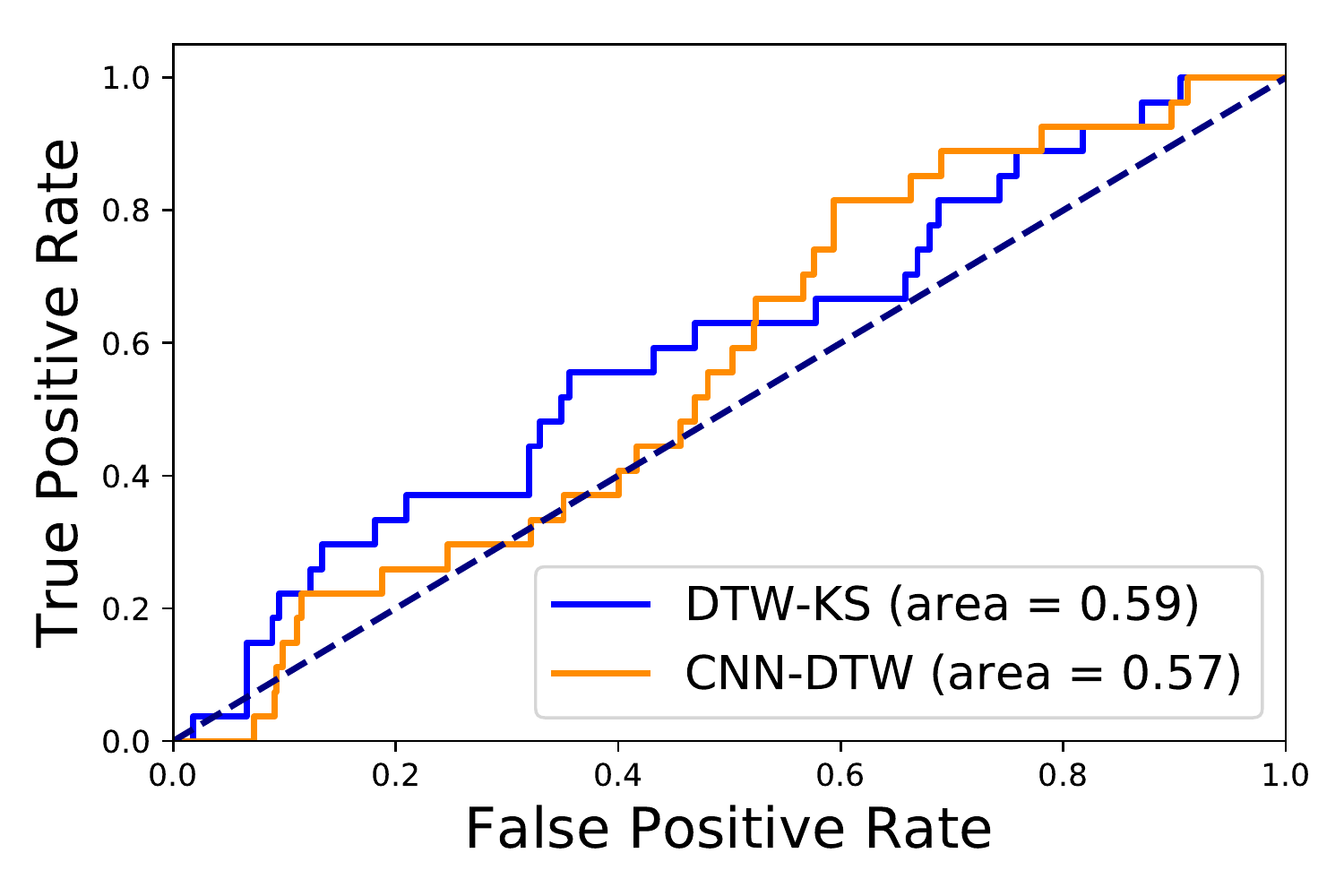} \label{fig:roc_d}}\\[-6pt]
\subfloat[Part 5][\vspace{-4pt}Keyword: War]{\includegraphics[width=1.5in]{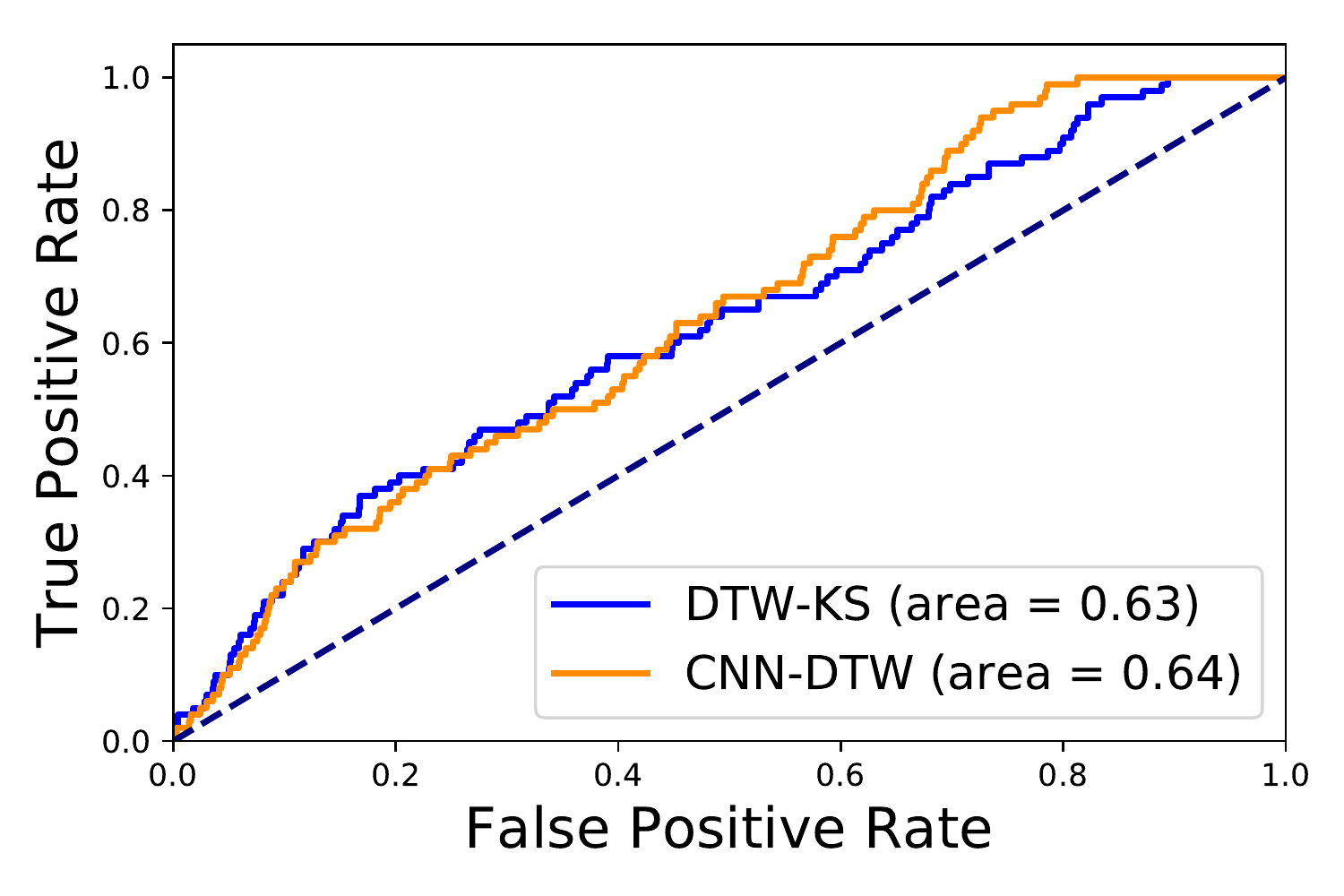} \label{fig:roc_e}}
\subfloat[Part 6][\vspace{-4pt}Keyword: Wounded]{\includegraphics[width=1.5in]{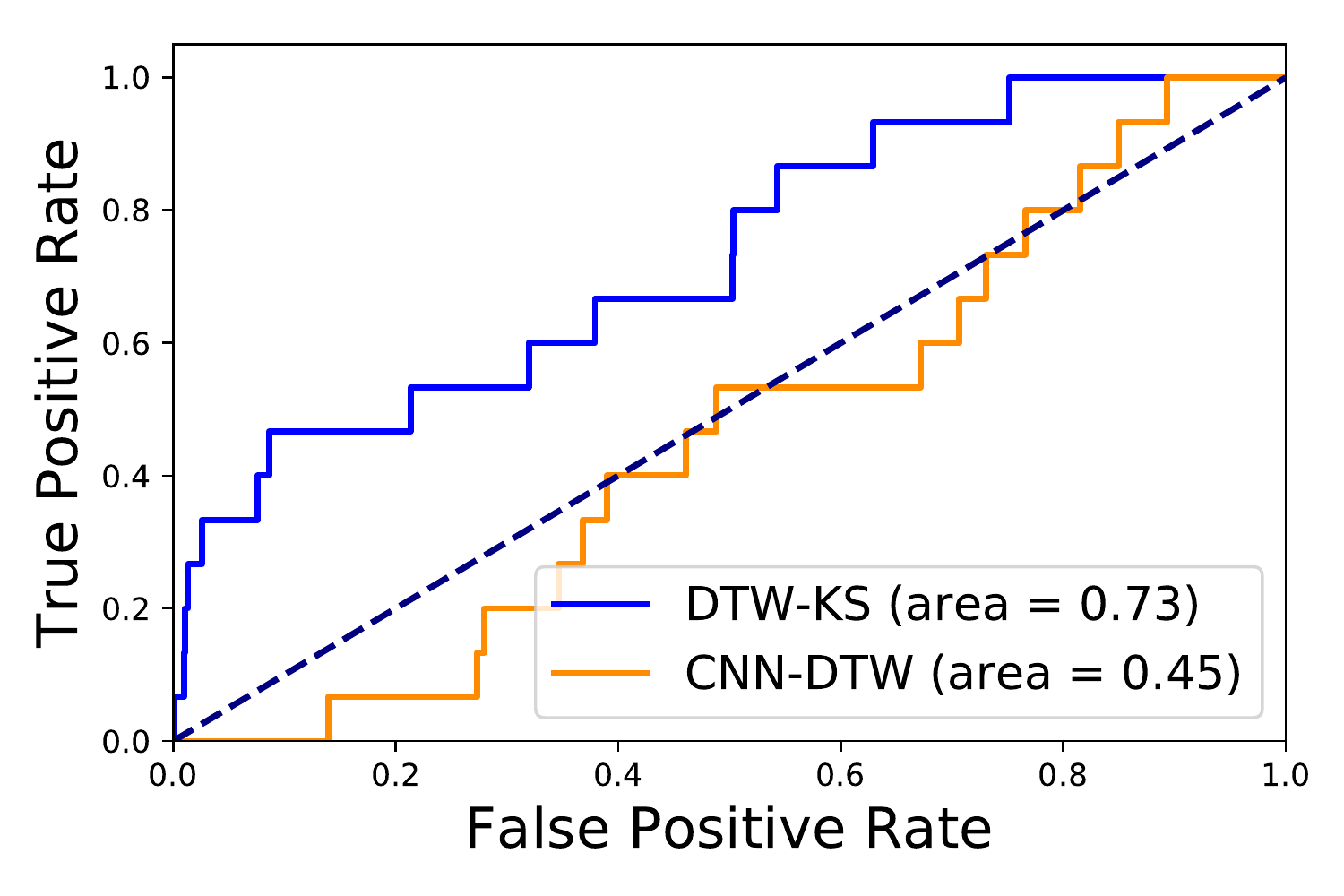} \label{fig:roc_f}}
% \vspace{-5pt}
\caption{Receiver operating characteristic plots for selected keywords for the DTW baseline and the proposed systems.}
\label{fig:ROC}
\vspace*{-10pt}
\end{figure}

In a qualitative analysis we found that performance differed significantly between different keyword types.% \todo{HK: Added this sentence.}
Table~\ref{Tab:roc} therefore analyses the development set performance of the CNN-DTW system on selected keywords. 
Examples (a), (c) and (e) are among the best-detected keywords, while (b), (d) and (f) are among the worst. 
A keyword specific threshold (at the EER) was used for this analysis.
%and is set to the threshold at the point of EER. 
As an example, Table~\ref{Tab:roc}(a) shows that for the keyword {\it{Government}}, at the EER, 93 of the 156 occurrences are correctly detected and 1683 of the 2832 negatives have also been correctly classified. 
As described in Section~\ref{Sec_Radio}, we intend to incorporate a human analyst in our overall radio-browsing approach; here we picked a threshold at the EER for the purpose of analysis, but a different threshold could be used depending on how the human analyst would want to balance correct detections and false positives.
% If a human analyst was used in the loop of detecting keywords (as we use in our overall radio-browsing approach described in ), a lower detection threshold could potentially be used.
% In fact, even the keywords with lower AUC seem to be identified successfully several times.
% Bearing the presence of human analysts in mind, a lower detection threshold my be feasible.
%Lowering the threshold, considering the  in the loop, for the keyword under consideration a improved performance can be obtained. 
The ROC plots for each of the keywords in Table~\ref{Tab:roc}(a-f) are shown in Figures~\ref{fig:ROC}(a-f), and are compared to those of the DTW-KS system. 
We notice from the plots that the performance of the CNN-DTW system is closer to the DTW baseline for keywords occurring more often in the SABN training set (shown in brackets in Table~\ref{Tab:roc}) and for keywords with distinct pronunciations, such as {\it{HIV}}.
The distribution of the keywords in the SABN {{train}}, {{development}} and {{test}} sets are shown in Figure~\ref{fig:key_dist}, and we see that \textit{Government} and \textit{War} are among the most frequent words. Note, again, that we never use the transcriptions of these sets during training or validation.
% \textcolor{green}{The reason for low AUC for some of the keywords may be due to the low frequency of their occurrence in the training set. The distribution of the keywords in {{train}}, {{development}} and {{test}} sets are shown in Figure~\ref{fig:key_dist}.}  

In a practical setting, the computational complexity of running the keyword spotter on live audio is an extremely important consideration. %, since computational resources are limited.
In this regard, our CNN-DTW keyword spotter shows a clear advantage over its DTW-KS counterpart.
%Another issue that is considered is the execution time for testing the incoming audio. 
%The DTW being a method of comparison of the target word to the incoming audio is slow. 
As indicated in Table~\ref{Tab:Res1}, the application of the DTW baseline systems for all 40 keywords and over all utterances in the 10-hour test (Table~\ref{SABC_data}) set was approximately 15 hours on a 20-core machine.
%This does not suit our purpose which requires the keyword spotter to function near real-time. 
The CNN-DTW system, on the other hand, can process the same data in approximately 5 minutes on a conventional desktop PC with a single {{GeForce GTX 1080}} GPU. 
Thus, we are able to process the audio 180 times faster using the CNN-DTW system, making it highly attractive for the cost-effective continuous monitoring of live audio streams. %, despite the drop in scores relative to DTW-KS.
%on a single core system with a GPU than that of DTW which takes 15 hours on a 20-core machine.

\vspace*{-7.5pt}
\begin{table}[t]
\centering
\captionsetup{justification = centering}
\caption{Keyword spotting performance on development and test sets with the execution time on the test set in minutes.}
\vspace*{-7.5pt}
\label{Tab:Res1}
{
\renewcommand{\arraystretch}{1.2}
\eightpt
\begin{tabular}{|l|l|l|l|l|c|}
\hline
\multirow{2}{*}{}                                                         & \multicolumn{2}{c|}{\textbf{AUC}}                                                        & \multicolumn{2}{c|}{\textbf{EER}}                                                        & \multirow{2}{*}{\textbf{\begin{tabular}[c]{@{}c@{}}Time\\ (min)\end{tabular}}} \\ \cline{2-5}
                                                                          & \multicolumn{1}{c|}{\textit{\textbf{dev}}} & \multicolumn{1}{c|}{\textit{\textbf{test}}} & \multicolumn{1}{c|}{\textit{\textbf{dev}}} & \multicolumn{1}{c|}{\textit{\textbf{test}}} &                                                                                 \\ \hline \hline
\textbf{CNN}                                                              & 0.5698                                     & 0.5448                                      & 0.4435                                     & 0.4771                                      & 55                                                                              \\ %\hline
\textbf{\begin{tabular}[c]{@{}l@{}}DTW-QbyE\end{tabular}}            & 0.6639                                     & 0.6612                                      & 0.3864                                     & 0.3885                                      & 900                                                                            \\ %\hline
\textbf{\begin{tabular}[c]{@{}l@{}}DTW-KS\end{tabular}}               & 0.7556                                     & 0.7515                                      & 0.3092                                     & 0.3162                                      & 900                                                                            \\ %\hline
\textbf{\begin{tabular}[c]{@{}l@{}}CNN-DTW\end{tabular}}               & 0.636                                      & 0.6285                                      & 0.4073                                     & 0.4161                                      & 5                                                                               \\ %\hline
\textbf{\begin{tabular}[c]{@{}l@{}}CNN-DTW\\ inc.\ GNL\end{tabular}} & \multicolumn{1}{c|}{0.6443}                & 0.6357                                      & 0.4036                                     & 0.4092                                      & 5                                                                               \\ \hline
\end{tabular}}
\vspace*{-10pt}
\end{table}

\begin{table}[]
\centering
\captionsetup{justification = centering}
\caption{Analysis of the 3 best performing and the 3 worst performing keywords. The number of occurrences of each keyword in the SABN corpus is shown in brackets. The absolute number of true positives (TP), false positives (FP), true negatives (TN) and false negatives (FN) are shown.}
\vspace*{-8.5pt}
\label{Tab:roc}
{
\renewcommand{\arraystretch}{1.2}
\eightpt
\begin{tabular}{cclcc}
\multicolumn{2}{c}{\textbf{Government (156)}}                                                                                                     &                       & \multicolumn{2}{c}{\textbf{Attack (51)}}                                                                                                        \\ \cline{1-2} \cline{4-5} 
\multicolumn{1}{|c|}{\begin{tabular}[c]{@{}c@{}}TP:  93\end{tabular}} & \multicolumn{1}{c|}{\begin{tabular}[c]{@{}c@{}}FP: 1149\end{tabular}} & \multicolumn{1}{l|}{} & \multicolumn{1}{c|}{\begin{tabular}[c]{@{}c@{}}TP: 25\end{tabular}} & \multicolumn{1}{c|}{\begin{tabular}[c]{@{}c@{}}FP: 1481\end{tabular}} \\ \cline{1-2} \cline{4-5} 
\multicolumn{1}{|c|}{\begin{tabular}[c]{@{}c@{}}FN: 63\end{tabular}}  & \multicolumn{1}{c|}{\begin{tabular}[c]{@{}c@{}}TN: 1683\end{tabular}} & \multicolumn{1}{l|}{} & \multicolumn{1}{c|}{\begin{tabular}[c]{@{}c@{}}FN: 26\end{tabular}} & \multicolumn{1}{c|}{\begin{tabular}[c]{@{}c@{}}TN: 1456\end{tabular}} \\ \cline{1-2} \cline{4-5} 
\multicolumn{2}{c}{(a)}                                                                                                                            &                       & \multicolumn{2}{c}{(b)}                                                                                                                          \\
\multicolumn{2}{c}{\textbf{HIV (21)}}                                                                                                              &                       & \multicolumn{2}{c}{\textbf{Health (27)}}                                                                                                         \\ \cline{1-2} \cline{4-5} 
\multicolumn{1}{|c|}{\begin{tabular}[c]{@{}c@{}}TP: 14\end{tabular}}  & \multicolumn{1}{c|}{\begin{tabular}[c]{@{}c@{}}FP: 1032\end{tabular}} & \multicolumn{1}{l|}{} & \multicolumn{1}{c|}{\begin{tabular}[c]{@{}c@{}}TP: 14\end{tabular}} & \multicolumn{1}{c|}{\begin{tabular}[c]{@{}c@{}}FP: 1422\end{tabular}} \\ \cline{1-2} \cline{4-5} 
\multicolumn{1}{|c|}{\begin{tabular}[c]{@{}c@{}}FN: 7\end{tabular}}   & \multicolumn{1}{c|}{\begin{tabular}[c]{@{}c@{}}TN: 1935\end{tabular}} & \multicolumn{1}{l|}{} & \multicolumn{1}{c|}{\begin{tabular}[c]{@{}c@{}}FN: 13\end{tabular}} & \multicolumn{1}{c|}{\begin{tabular}[c]{@{}c@{}}TN: 1539\end{tabular}} \\ \cline{1-2} \cline{4-5} 
\multicolumn{2}{c}{(c)}                                                                                                                            &                       & \multicolumn{2}{c}{(d)}                                                                                                                          \\
\multicolumn{2}{c}{\textbf{War (100)}}                                                                                                             &                       & \multicolumn{2}{c}{\textbf{Wounded (15)}}                                                                                                       \\ \cline{1-2} \cline{4-5} 
\multicolumn{1}{|c|}{\begin{tabular}[c]{@{}c@{}}TP: 58\end{tabular}}  & \multicolumn{1}{c|}{\begin{tabular}[c]{@{}c@{}}FP: 1222\end{tabular}} & \multicolumn{1}{l|}{} & \multicolumn{1}{c|}{\begin{tabular}[c]{@{}c@{}}TP: 8\end{tabular}}  & \multicolumn{1}{c|}{\begin{tabular}[c]{@{}c@{}}FP: 1452\end{tabular}}  \\ \cline{1-2} \cline{4-5} 
\multicolumn{1}{|c|}{\begin{tabular}[c]{@{}c@{}}FN: 42\end{tabular}}   & \multicolumn{1}{c|}{\begin{tabular}[c]{@{}c@{}}TN: 1666\end{tabular}} & \multicolumn{1}{l|}{} & \multicolumn{1}{c|}{\begin{tabular}[c]{@{}c@{}}FN: 7\end{tabular}}   & \multicolumn{1}{c|}{\begin{tabular}[c]{@{}c@{}}TN: 1521\end{tabular}}  \\ \cline{1-2} \cline{4-5} 
\multicolumn{2}{c}{(e)}                                                                                                                            &                       & \multicolumn{2}{c}{(f)}                                                                                                                         
\end{tabular} }
 \vspace{-10pt}
\end{table}

\section{Conclusions}
\label{Sec:CONclusion}
We have shown that, by combining CNNs and DTW, it is possible to obtain an ASR-free keyword spotting system that is fast enough for real-time processing. 
Because it requires only a small set of easily-obtained isolated keywords for training, it is suitable for a low-resource keyword detection scenario in which no further transcribed speech is available.
%and the only resource available is independently recorded isolated words that has to be identified. 
%It has also been shown that the system performs reasonably well when these target words are independently recorded. 
The performance of the proposed system is comparable with the DTW-based query-by-example (QbyE) approaches which is commonly employed in these situations, achieving an AUC of 0.64.    %$63.6\%$. 
%It was noted that the performance of the system improves by  $1\%$ when a gaussian noise layer is added between the input and the convolution layers with a ROC-AUC of $64.43\%$ as the noise layer aids in generalization. 
%It remains to be seen in ongoing work whether augmenting the training data using added gaussian noise improves the result further.
Besides requiring minimal resources to train, the proposed system is also computationally much more efficient than its DTW counterpart, and is suitable for the real-time application required by the United Nation's ongoing efforts in humanitarian monitoring.
 
\section{Acknowledgements}
We thank the NVIDIA corporation for the donation of GPU equipment used for this research. We also gratefully acknowledge the support of Telkom South Africa.

\newpage
\bibliographystyle{IEEEtran}

\bibliography{mybib}

\begin{thebibliography}{10}
\providecommand{\url}[1]{#1}
\csname url@samestyle\endcsname
\providecommand{\newblock}{\relax}
\providecommand{\bibinfo}[2]{#2}
\providecommand{\BIBentrySTDinterwordspacing}{\spaceskip=0pt\relax}
\providecommand{\BIBentryALTinterwordstretchfactor}{4}
\providecommand{\BIBentryALTinterwordspacing}{\spaceskip=\fontdimen2\font plus
\BIBentryALTinterwordstretchfactor\fontdimen3\font minus
  \fontdimen4\font\relax}
\providecommand{\BIBforeignlanguage}[2]{{%
\expandafter\ifx\csname l@#1\endcsname\relax
\typeout{** WARNING: IEEEtran.bst: No hyphenation pattern has been}%
\typeout{** loaded for the language `#1'. Using the pattern for}%
\typeout{** the default language instead.}%
\else
\language=\csname l@#1\endcsname
\fi
#2}}
\providecommand{\BIBdecl}{\relax}
\BIBdecl

\bibitem{Vosoughi_ICDMW15}
S.~Vosoughi and D.~Roy, ``A human-machine collaborative system for identifying
  rumors on {T}witter,'' in \emph{Proc. ICDMW}, 2015.

\bibitem{Wegrzyn_CASoN11}
K.~Wegrzyn-Wolska, L.~Bougueroua, and G.~Dziczkowski, ``Social media analysis
  for e-health and medical purposes,'' in \emph{Proc. CASoN}, 2011.

\bibitem{Burnap15}
P.~Burnap, G.~Colombo, and J.~Scourfield, ``Machine classification and analysis
  of suicide related communication on {T}witter,'' in \emph{Proc. ACM-HT},
  2015.

\bibitem{Menon2017}
R.~Menon, A.~Saeb, H.~Cameron, W.~Kibira, J.~Quinn, and T.~Niesler,
  ``Radio-browsing for developmental monitoring in {U}ganda,'' in \emph{Proc.
  ICASSP}, 2017.

\bibitem{Saeb2017}
A.~Saeb, R.~Menon, H.~Cameron, W.~Kibira, J.~Quinn, and T.~Niesler, ``Very low
  resource radio browsing for agile developmental and humanitarian
  monitoring,'' in \emph{Proc. INTERSPEECH}, 2017.

\bibitem{Sainath2015}
T.~N. Sainath, O.~Vinyals, A.~Senior, and H.~Sak, ``Convolutional, long
  short-term memory, fully connected deep neural networks recognition,'' in
  \emph{Proc. ICASSP}, 2015.

\bibitem{Zhang2017}
Y.~Zhang, W.~Chan, and N.~Jaitly, ``Very deep convolutional networks for
  end-to-end speech recognition,'' in \emph{Proc. ICASSP}, 2017.

\bibitem{Larson12}
M.~Larson and G.~J.~F. Jones, ``Spoken content retrieval: A survey of
  techniques and technologies,'' \emph{Foundations and Trends in Information
  Retrieval}, vol.~5, no. 4-5, pp. 235--422, 2012.

\bibitem{Mandal14}
A.~Mandal, K.~R.~P. Kumar, and P.~Mitra, ``Recent developments in spoken term
  detection: A survey,'' \emph{International Jour. of Speech Technology},
  vol.~17, no.~2, pp. 183--198, 2014.

\bibitem{Hazen2009}
T.~J. Hazen, W.~Shen, and C.~White, ``Query-by-example spoken term detection
  using phonetic posteriorgram templates,'' in \emph{Proc. ASRU}, 2009.

\bibitem{Zhang2009}
Y.~Zhang and J.~R. Glass, ``Unsupervised spoken keyword spotting via segmental
  dtw on gaussian posteriorgrams,'' in \emph{Proc. ASRU}, 2009.

\bibitem{Levin2013}
K.~Levin, K.~Henry, A.~Jansen, and K.~Livescu, ``Fixed-dimensional acoustic
  embeddings of variable-length segments in low-resource settings,'' in
  \emph{Proc. ASRU}, 2013.

\bibitem{Chung2016}
Y.~Chung, C.~Wu, C.~Shen, H.~Lee, and L.~Lee, ``Audio word2vec: Unsupervised
  learning of audio segment representations using sequence-to-sequence
  autoencoder,'' in \emph{Proc. INTERSPEECH}, 2016.

\bibitem{Chen2015}
G.~Chen, C.~Parada, and T.~N. Sainath, ``Query-by-example keyword spotting
  using long short-term memory networks,'' in \emph{Proc. ICASSP}, 2015.

\bibitem{settle+livescu_slt16}
S.~Settle and K.~Livescu, ``Discriminative acoustic word embeddings: Recurrent
  neural network-based approaches,'' in \emph{Proc. SLT}, 2016.

\bibitem{Kamper2016}
H.~Kamper, W.~Wang, and K.~Livescu, ``Deep convolutional acoustic word
  embeddings using word-pair side information,'' in \emph{Proc. ICASSP}, 2016.

\bibitem{Audhkhasi2017}
K.~Audhkhasi, A.~Rosenberg, A.~Sethy, B.~Ramabhadran, and B.~Kingsbury,
  ``End-to-end {ASR}-free keyword search from speech,'' in \emph{Proc. ICASSP},
  2017.

\bibitem{Kamper2015}
H.~Kamper, F.~D. Wet, T.~Hain, and T.~Niesler, ``Capitalising on {N}orth
  {A}merican speech resources for the development of a {S}outh {A}frican
  {E}nglish large vocabulary speech recognition system,'' \emph{Computer Speech
  and Language}, vol.~28, no.~6, pp. 1255--1268, 2014.

\bibitem{park+glass_taslp08}
A.~S. Park and J.~R. Glass, ``Unsupervised pattern discovery in speech,''
  \emph{IEEE Trans. Audio, Speech, Language Process.}, vol.~16, no.~1, pp.
  186--197, 2008.

\bibitem{jansen+vandurme_interspeech12}
A.~Jansen and B.~Van~Durme, ``Indexing raw acoustic features for scalable zero
  resource search,'' in \emph{Proc. Interspeech}, 2012.

\bibitem{palaz+etal_interspeech16}
D.~Palaz, G.~Synnaeve, and R.~Collobert, ``Jointly learning to locate and
  classify words using convolutional networks,'' in \emph{Proc. Interspeech},
  2016.

\bibitem{SainathPara2015}
T.~N. Sainath and C.~Parada, ``Convolutional neural networks for
  small-footprint keyword spotting,'' in \emph{Proc. INTERSPEECH}, 2015.

\bibitem{xu2015empirical}
B.~Xu, N.~Wang, T.~Chen, and M.~Li, ``Empirical evaluation of rectified
  activations in convolutional network,'' \emph{arXiv preprint
  arXiv:1505.00853}, 2015.

\bibitem{kingma2014adam}
D.~P. Kingma and J.~Ba, ``Adam: A method for stochastic optimization,''
  \emph{arXiv preprint arXiv:1412.6980}, 2014.

\end{thebibliography}

\end{document}